\documentclass{article}
\usepackage{ijcai22}

\usepackage{times}

\usepackage{soul}
\usepackage{url}
\usepackage[hidelinks]{hyperref}
\usepackage[utf8]{inputenc}
\usepackage[small]{caption}
\usepackage{graphicx}
\usepackage{amsmath}
\usepackage{booktabs}
\usepackage{algorithm}
\usepackage{algorithmic}
\usepackage{nicefrac}

\newcommand{\citet}[1]{\citeauthor{#1}~\shortcite{#1}}
\newcommand{\citep}{\cite}

\usepackage{amsmath}
\usepackage{amssymb}
\usepackage[small]{subfigure}
\usepackage{bbm}

\usepackage{blkarray, bigstrut}
\usepackage{enumitem}
\setlist[enumerate,itemize]{itemsep=0pt,topsep=0pt}

\usepackage[textsize=small,textwidth=3cm]{todonotes}
\definecolor{kentuckyblue}{RGB}{0, 93, 170}

\newtheorem{definition}{Definition}

\newcommand{\reals}{\ensuremath{\mathbb{R}}}
\newcommand{\mdp}{\ensuremath{\mathcal{M}}}
\newcommand{\nominalmdp}{\ensuremath{\mathcal{M}^{\mathcal{N}}}}
\newcommand{\trueconstrainedmdp}{\ensuremath{\mathcal{M}^{\mathcal{C}^*}}}
\newcommand{\learnedmdp}{\ensuremath{\mathcal{M}^{\mathcal{C}}}}
\newcommand{\states}{\ensuremath{\mathcal{S}}}
\newcommand{\actions}{\ensuremath{\mathcal{A}}}
\newcommand{\demonstrations}{\ensuremath{\mathcal{D}}}
\newcommand{\constraints}{\ensuremath{\mathcal{C}}}

\newcommand{\optnom}{\ensuremath{\pi_{\mathcal{N}}^*}}
\newcommand{\optconstrained}{\ensuremath{\pi_{\mathcal{C}^*}^*}}

\setlist[enumerate,itemize]{itemsep=0.5pt,topsep=0pt}
\title{Learning Behavioral Soft Constraints from Demonstrations}

\author{
Arie Glazier $^1$, 
Andrea Loreggia $^2$, 
Nicholas Mattei $^1$, 
Taher Rahgooy $^3$, \\
Francesca Rossi $^4$, \And
Brent Venable $^3$
\affiliations
$^1$Tulane University, New Orleans, LA, USA\\
$^2$University of Brescia, Italy, \\
$^3$University of West Florida and IHMC, Pensacola, FL, USA \\
$^4$IBM Research, Yorktown Heights, NY, USA
\emails
{\small adglazier@gmail.com, andrea.loreggia@gmail.com, nsmattei@tulane.edu, trahgooy@students.uwf.edu, Francesca.Rossi2@ibm.com, bvenable@ihmc.org}
}

\begin{document}

\maketitle
\begin{abstract}
Many real-life scenarios require humans to make difficult trade-offs: do we always follow all the traffic rules or do we violate the speed limit in an emergency? These scenarios force us to evaluate the trade-off between collective rules and norms with our own personal objectives and desires.
To create effective AI-human teams, we must equip
AI agents with a model of how humans make these trade-offs in complex environments when there are implicit and explicit rules and constraints. Agent equipped with these models will be able to mirror human behavior and/or to draw human attention to situations where decision making could be improved.
To this end, we propose a novel inverse reinforcement learning (IRL) method: Max Entropy Inverse Soft Constraint IRL (MESC-IRL), for learning implicit hard and soft constraints over states, actions, and state features from demonstrations in deterministic and non-deterministic environments modeled as Markov Decision Processes (MDPs). Our method enables agents implicitly learn human constraints and desires without the need for explicit modeling by the agent designer and to transfer these constraints between environments. Our novel method generalizes prior work which only considered deterministic hard constraints and achieves state of the art performance.

\end{abstract}

\section{Introduction}

Implicit and explicit constraints, typically arising from morals, norms, or rules, are present in many decision making scenarios, and their presences forces us to make difficult decisions: do we always satisfy all constraints, or do we violate some of them in exceptional circumstances? Many techniques can be used to combine constraints and goals so that an autonomous agent rationally minimizes constraint violations while achieving the given goal \cite{noothigattu-2019-ethicalvalues}. 
Often these constraints are not hard but rather soft and violating them can be considered as incurring in a cost.
Moreover, these constraints are often not explicitly given, but need to be inferred from observations of how other agents act within a constrained environment. Learning constraints from demonstrations is an important topic in the domains of inverse reinforcement learning \cite{scobee-2020-maximumlikelihood,abbeel2004apprenticeship}, which is used to implement AI safety goals including value alignment \cite{russell2015research,balakrishnan2018incorporating,LoMaRoVe18} and to circumvent reward hacking \cite{amodei2016concrete,ray2019benchmarking}. Recent work has focused on building ethically bounded agents \cite{balakrishnan2018incorporating} that comply with ethical or moral theories of action \cite{svegliato2021ethically,FoMa19a}. Following the work of \citet{scobee-2020-maximumlikelihood}, we propose an architecture that, given access to a model of the environment and to demonstrations of constrained behavior, is able to learn constraints over states, actions, or state features. Our method, MESC-IRL, performs comparably with the state of the art and is more general, as it can handle both hard and soft constraints in both deterministic and non-deterministic environments. It is also decomposable into features of the environment, supporting the transfer of the learned constraints to new settings.

\smallskip
\noindent
\textbf{Contributions.}
We propose and evaluate a novel method, MESC-IRL, that is able to learn both hard and soft constraints over states, actions, and state features in both deterministic and non-deterministic MDPs from a set of demonstrations. This method strictly generalizes existing methods in the literature and achieves state of the art performance in our testing in gridworld domains. Our method is also decomposable into features of the environment, which supports transferring learned constraints between environments.

\begin{figure}[ht]
\centering
\begin{minipage}[t]{0.7\linewidth}
  \centering
  \includegraphics[width=\linewidth]{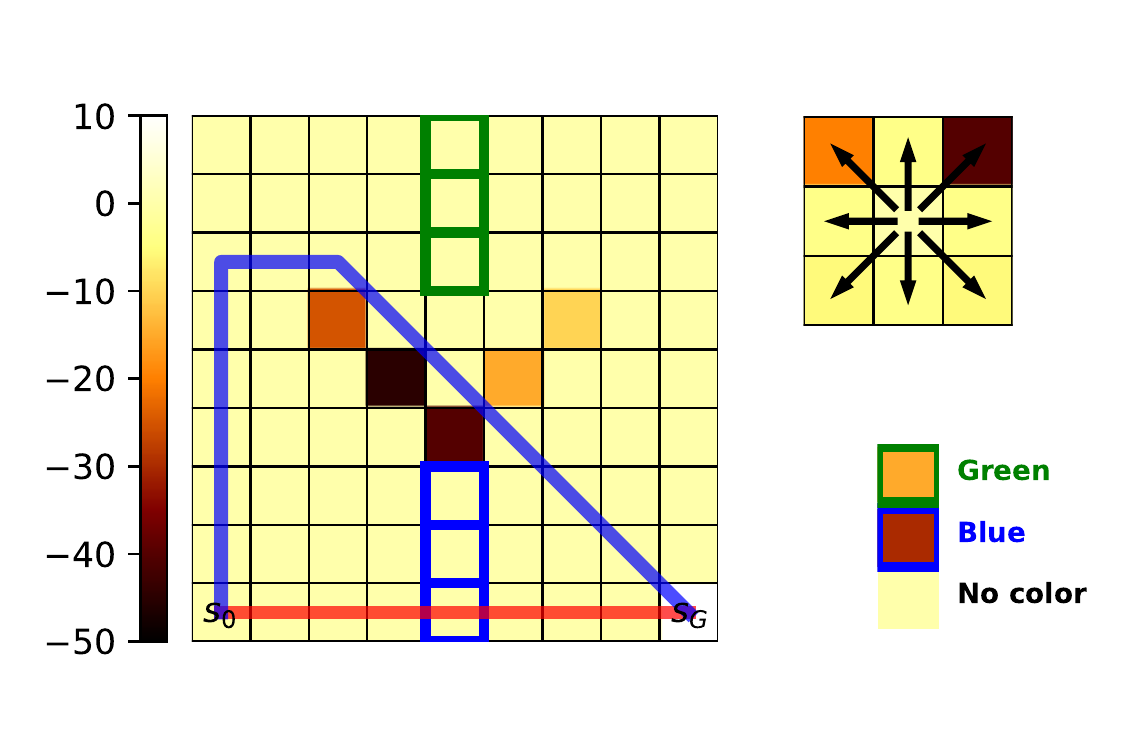}
\end{minipage}%
\caption{Example grid with constraints of varying costs over actions, state occupancy, and state features. Note $\optnom$ for $\nominalmdp$ is the red trajectory, obtained by an agent that does not know the constraints while $\optconstrained$ for $\trueconstrainedmdp$ is the blue trajectory.} 
\label{fig:example_grid}
\end{figure}

\section{Preliminaries and Related Work}\label{sec:prelim}

We first provide the preliminary notions on the context of our work, that is, constrained Markov Decision Processes and Reinforcement Learning \cite{sutton-barto-rl}. We then review fundamental concepts and methods on Inverse Reinforcement Learning \cite{Ng2000,abbeel2004apprenticeship} and background on Constrained Markov Decision Processes \cite{altman-1999-constrained} including related work on learning constraints \cite{ziebart-2008-maxentropy,anwar-2021-icrl,scobee-2020-maximumlikelihood}, which we will leverage to develop our novel method for learning soft constraints \cite{rossi2006handbook} from demonstrations \cite{chou2018learning}. 

\subsection{Markov Decision Processes and Reinforcement Learning}

A finite-horizon Markov Decision Process (MDP) $\mdp$ is a model for sequential decision making over a number of time steps $t \in T$ defined by a tuple $(\states, \actions, P, D_0, \phi, \gamma, R)$ \cite{sutton-barto-rl}. $\states$ is a finite set of discrete states; $\actions$ is a set of actions where, for every state $s \in \states$ we may only have the ability to take a subset of the whole action set, $\{\actions_s\} \subseteq \actions$; $P: \states \times \actions \times \states \rightarrow [0,1] $ is a model of the environment given as transition probabilities where $P(s_{t+1}|s_t,a_t)$ is the probability of transitioning to state $s_{t+1}$ from state $s_t$ after taking action $a_t \in \{\actions_{s_t}\}$ at time $t$. $D_0: \states \rightarrow [0,1]$ is a distribution over start states;  $\phi : \states \times \actions \times \states \rightarrow \reals^k$ is a mapping from the transitions to a $k$-dimensional space of features; $\gamma \in [0,1)$ is a discount factor; and $R: \states \times  \actions \times \states \rightarrow \reals$ is a scalar reward received by the agent for being in one state and transitioning to another state at time $t$, written as $R(s_t,a_t,s_{t+1})$. 

An agent acts within the environment defined by the MDP, generating a sequence of actions called a \emph{trajectory} of length $t$. Let $\tau = ((s_1,a_1, s_{2}),...,(s_{t-1},a_{t-1}, s_{t})) \in (\states \times \actions \times \states)^t$. We evaluate the quality of a particular trajectory in terms of the amount of reward accrued over the trajectory, subject to discounting. Formally, $R(\tau) = \sum^t_{i=1} \gamma^i R(s_i, a_i, s_{i+1})$. A policy, $\pi: \states \rightarrow \mathcal{P}(\mathcal{A})$ is a mapping of probability distributions over actions to every state $s$ such that $\pi(s,a)$ is the probability of taking action $a$ in state $s$. We can also write the probability of a trajectory $\tau$ under a policy as $\pi(\tau) = \prod_{(s_t, a_t, s_{t+1})} a_t$. The feature vector associated with trajectory $\tau$ is defined as the summation over all transition feature vectors in $\tau$, $\phi(\tau) = \sum_{(s_t, a_t, s_{t+1}) \in \tau} \phi(s_t, a_t, s_{t+1})$

The goal within an MDP is to find a policy $\pi^*$ that maximizes the expected reward, $J(\pi) = \mathbb{E}_{\tau \sim \pi}[R(\tau)]$ \cite{anwar-2021-icrl}. In the MDP literature, classical tabular methods are used to find $\pi^*$ including value iteration (VI) and Q-learning. Such method finds an optimal policy by estimating the expected reward for taking an action $a$ in a given state $s$, i.e., the $Q$-value of pair $(s,a)$, written $q(s,a)$. \cite{sutton-barto-rl}.

\subsection{Constrained MDPs and Inverse Reinforcement Learning}

We are interested in learning constraints from a set of demonstrations $\demonstrations$. Our goal is to create agents that can be trained to follow constraints that are not explicitly prohibited in the MDP, but should be avoided \cite{FoMa19a}. \citet{scobee-2020-maximumlikelihood} discusses the importance of such constraints: an MDP $\mdp$ may encode everything necessary about driving a car, e.g. the dynamics of steering and movements, but often one wants to add additional general constraints such as \emph{avoid obstacles on the way to the goal}. These constraints may be temporally complex and require many small constraints over states and actions, hence engineering a reward function that encodes these constraints may be a difficult or impossible task \cite{vazquez-2018-taskfromdemonstrations}.

One approach for learning constraints from demonstrations is to use techniques from inverse reinforcement learning (IRL): given a set of demonstrated trajectories $\mathcal{D}$ of an agent in an environment $\mdp$ with an unknown reward function $\mathcal{M} \setminus R$, IRL provides a set of techniques for learning a reward function $\hat{R}$ that explains the agent's demonstrated behavior \cite{abbeel2004apprenticeship,Ng2000}. However, this technique has many drawbacks: often there are many reward functions that lead to the same behavior \cite{scobee-2020-maximumlikelihood}, the reward functions may not be interpretable \cite{vazquez-2018-taskfromdemonstrations}, and there may be issues such as reward hacking -- wherein the agent learns to behave in ways that create reward but are not intended by the designer -- an important topic in the field of AI safety \cite{amodei2016concrete,ray2019benchmarking} and value alignment \cite{FoMa19a,russell2015research}. 

We follow the framework of \citet{altman-1999-constrained} and \citet{anwar-2021-icrl} and define a Constrained MDP $\learnedmdp$ which is a nominal MDP $\nominalmdp$ with an additional cost function $\constraints: \states \times \actions \times \states \rightarrow \reals$ and a budget $\alpha \geq 0$. We can then define the cost of a trajectory to be $c(\tau) = \sum^t_{i=1} c(s_i, a_i, s_{i+1})$. Setting $\alpha = 0$ is enforcing \emph{hard constraints}, i.e., we must never trigger constrained transitions. In this work, unlike the work of both \citet{scobee-2020-maximumlikelihood} and \citet{anwar-2021-icrl}, we are interested in learning \emph{soft constraints} \cite{rossi2006handbook}. Under a soft constraints paradigm, each constraint comes with a real-valued penalty/cost and the goal is to minimize the sum of penalties incurred by the agent.

Following \citet{scobee-2020-maximumlikelihood}, the task of constraint inference in IRL is defined as follows. Given a nominal MDP $\nominalmdp$ and a set of demonstrations $\mathcal{D}$ in ground-truth constrained world $\trueconstrainedmdp$, we wish to find the most likely set of constraints $\constraints$ that approximates the ground truth constraint set $\constraints^*$ that could modify $\nominalmdp$ to explain the demonstrations. We are concerned with three types of constraints:
\begin{description}
    \item \textbf{Action Constraints:} We may not want an agent to ever perform some (set of) action $a_i$;
    \item \textbf{Occupancy Constraints:} We may not want an agent to occupy a (set of) states $s_i$;
    \item \textbf{Feature Constraints:} Given a feature mapping transitions to a feature vector $\phi$, we may not want an agent to perform an action (or set of actions) in the presence of specific state features.
\end{description}
\noindent
Without loss of generality, we add the state and actions to the features. Hence, action and occupancy become specific cases of feature constraints.

To test our methods, we use the same grid world setup as \citet{scobee-2020-maximumlikelihood}. Within our grid world example, shown in Figure \ref{fig:example_grid} (Left), we have an action penalty of $-4$ for the cardinal directions, $-4 \times \sqrt{2}$ for taking the diagonal actions, and reaching the goal state has a reward of $10$. In Figure \ref{fig:example_grid} (Left) we set the constraint costs to various values but in all our experiments we fix the constraint costs on the generated grids for states, actions, and features to be $-50$. The feature vector $\phi$ is a one-hot vector representing the color associated with each state: blue, green, and no color. Throughout we assume a non-deterministic world with a $10\%$ chance of action failure, resulting in a random action.

\subsection{Related Work}

Building AI systems that adhere to human values and norms is a challenging task that has received significant attention in the literature in recent years. Our goal is to enable the construction of ethically bounded AI that acts as optimally as possible within the confines of our values and norms \cite{FoMa19a,LoMaRoVe18,balakrishnan2018incorporating}. Reinforcement learning has emerged as a popular paradigm for this task as it allows for modeling sequential decisions in complex environments where there may be competing values and goals \cite{abel2016reinforcement,BaBoMaRo18}. However, creating reward functions for these agents can be a complex task leading to issues of reward hacking \cite{amodei2016concrete,ray2019benchmarking} and value mis-specification \cite{russell2015research}, resulting in unintended or undesirable behavior by the agents. 

Various methods have been proposed to address these challenges including multi-objective MDPs \cite{rodriguez2021multi} and bounding the solution space of MDPs in order to ensure ethical compliance via constraint inference techniques \cite{svegliato2021ethically}. However, in both of these examples, the ethics or norms that one wants the agent to follow must be defined aprori as part of the overall reward function, and require expert knowledge of both the domain and the rules to be followed. We desire methods that are able to learn from human demonstrations as this allows a way to define ethics or norms that a system should follow that may not be encoded in the dynamics of the environment reward function in an easier way, without the need for expert engineering.

Within the work on learning ethics from demonstrations, \citet{noothigattu-2019-ethicalvalues} also build a system for learning norms from demonstrations in the domain of Pac-man. Their system is similar to ours but they employ a bandit based approach to predict which of two reward functions to follow, either a totally ethical reward function, defined via IRL, or the optimal policy for the nominal MDP. This method does not allow for extraction of the constraints and transfer between domains, and also does not necessarily jointly optimize both tasks. In other work \citet{wu2018low}, use a set of demonstrations and a per-state counting technique to ``shape'' a reward function with an additive term based on how often a particular action is taken in a set of demonstrations. The technique of \citet{wu2018low} is more limited than ours: they are only able to capture constraints on actions, and because they only use an empirical count, with no notion of minimality over constraints, as we use. Hence, their method is likely to find many constraints that are not real, leading to a large false positive rate. Additionally, the method of \citet{wu2018low} requires a large number of free parameters: minimum and maximum thresholds for actions as well as penalty parameters, that all must be hand tuned and hence require an agent designer to have fine grained knowledge of the target domain and desired behavior. Whereas in our model because we use learning techniques to find the minimal set of constraints necessary these parameters are all learned directly from the demonstrations.

Finally, within the topic of learning constraints from demonstrations the work of \citet{scobee-2020-maximumlikelihood} and \citet{anwar-2021-icrl} are the closest to the work presented here. In both \citet{scobee-2020-maximumlikelihood} and \citet{anwar-2021-icrl} the domain is restricted to deterministic MDPs, which we strictly generalize in this work. In the work of \citet{scobee-2020-maximumlikelihood} the definition of the cost function is limited to a set of state-actions $\states \times \actions$ and a deterministic setting, which allows them to define $\learnedmdp$ by substituting $\mathcal{A} = \{A_s\}$ with $\mathcal{A}^\mathcal{C} = \{A^\mathcal{C}_s\}$ in $\nominalmdp$. In order to extend this method to non-deterministic settings, we define the cost function over the set of transitions $\constraints: \states \times \actions \times \states \rightarrow \reals$ and derive our learning method by adapting techniques from \citet{ziebart-2010-causalentropy}.
In addition, both \citet{scobee-2020-maximumlikelihood} and \citet{anwar-2021-icrl} propose a greedy approach to infer a set of hard constraint $\constraints$ that explains the demonstrations $D$ on $\trueconstrainedmdp$, whereas in our method we use a learning approach based on maximum entropy to find the minimal set of constraints consistent with $\demonstrations$.
%


\begin{figure*}
\begin{minipage}[t]{.32\textwidth}
  \centering
  \includegraphics[width=\linewidth]{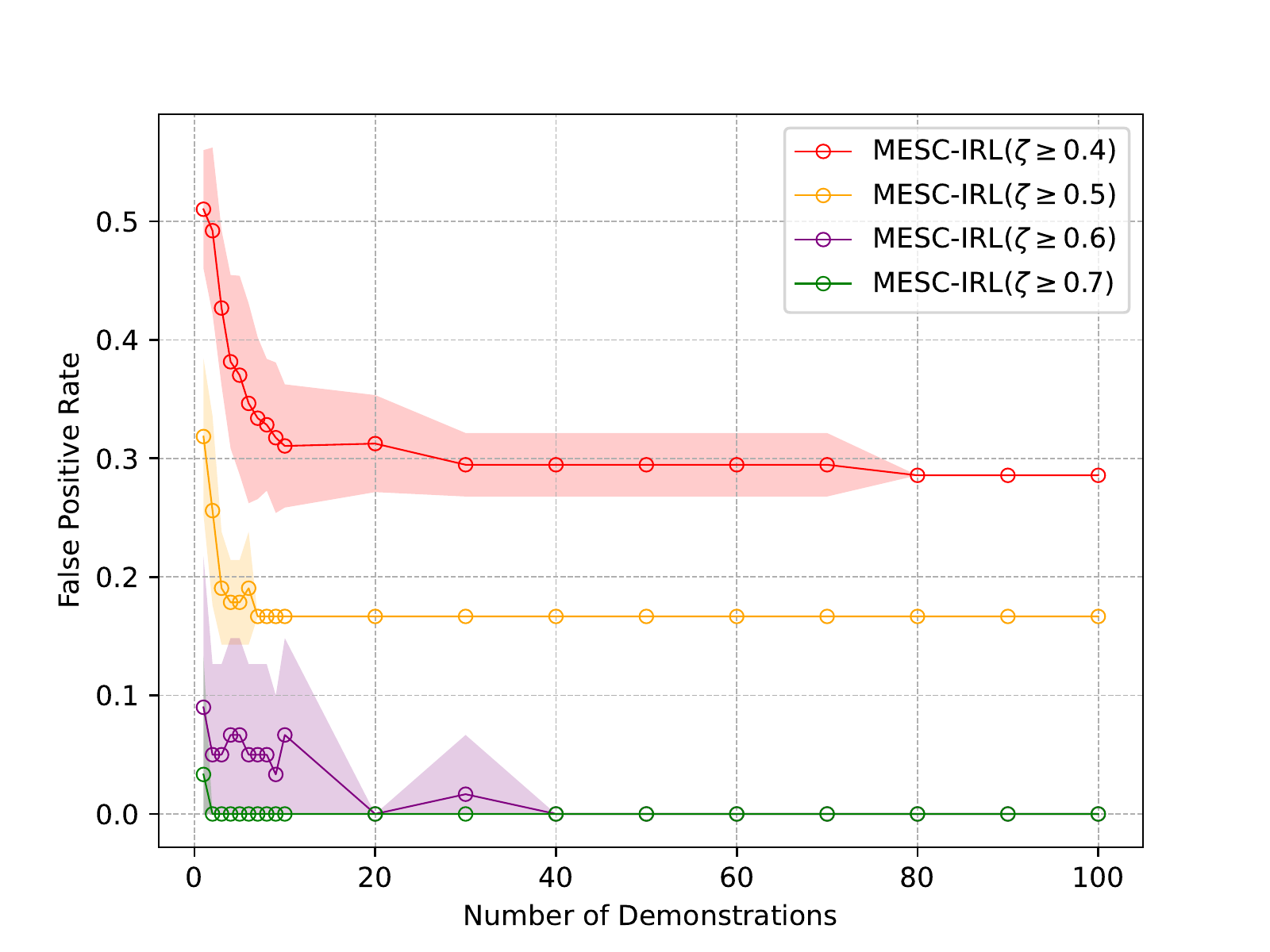}
\end{minipage}%
\hfill
\begin{minipage}[t]{.32\textwidth}
  \centering
  \includegraphics[width=\linewidth]{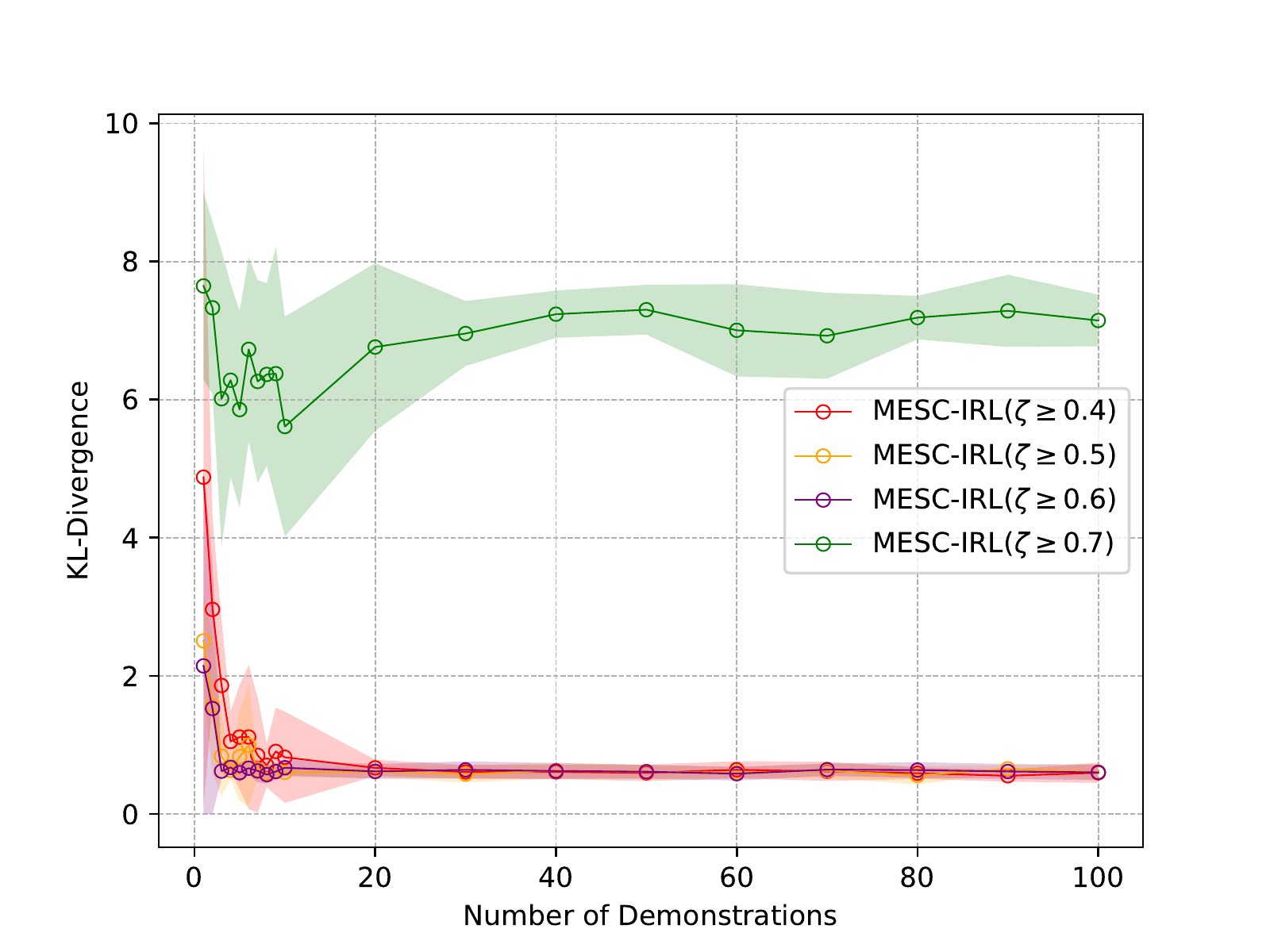}
\end{minipage}
\hfill
\begin{minipage}[t]{.32\textwidth}
  \centering
  \includegraphics[width=\linewidth]{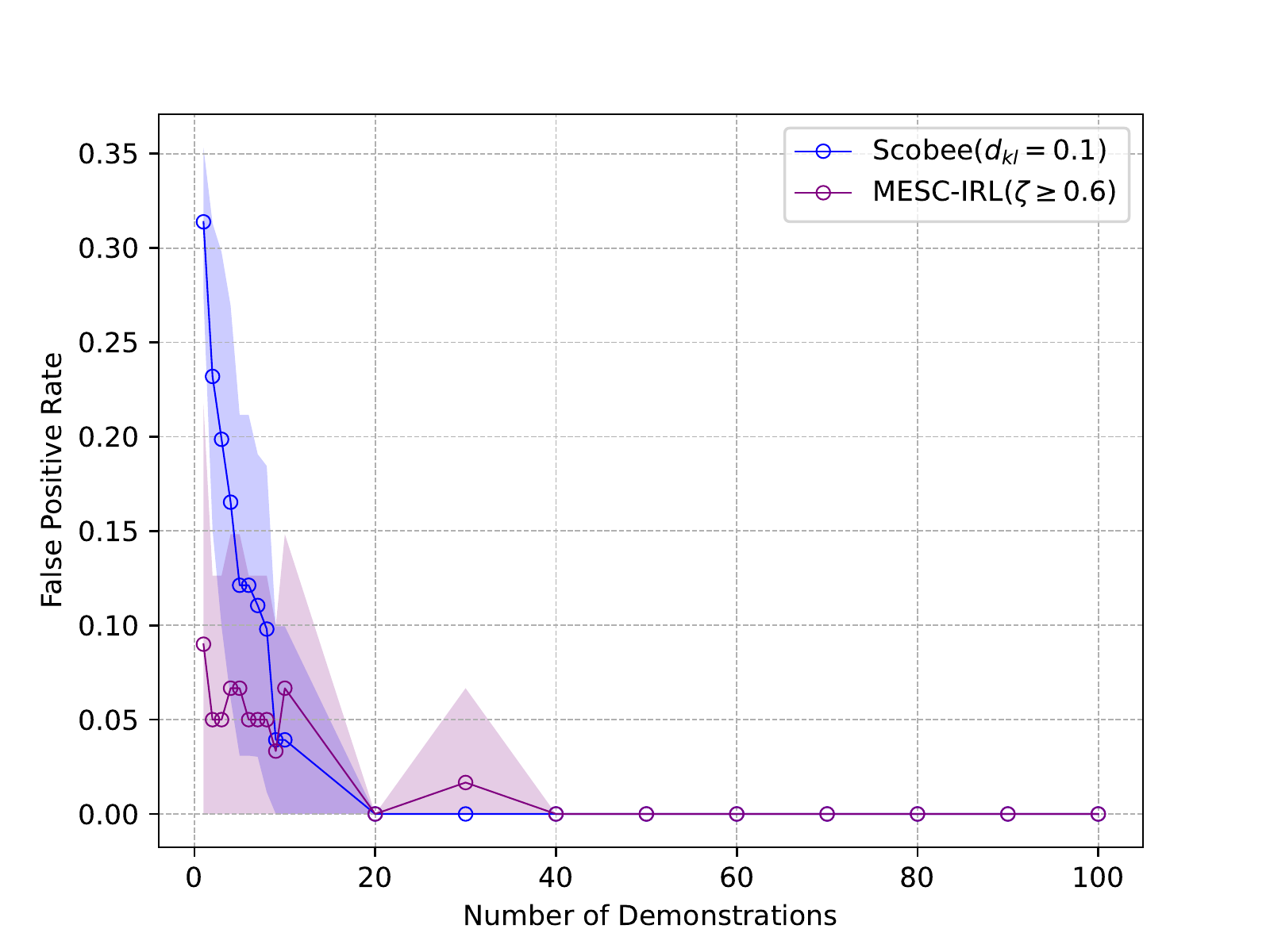}
\end{minipage}
\caption{Performance of MESC-IRL for various settings of $\zeta$ at recovering hard constraints in a deterministic setting according to false positive rate (left) and KL-Divergence from the demonstrations $\demonstrations$ (center), and a comparision with the best performing method of Scobee et al. (right) as we vary the number of demonstrations. Each point is the mean of 10 independent draws.}
%
\label{fig:stepping}
\end{figure*}

\section{MESC-IRL: Max Entropy Inverse Soft-Constraint Inverse Reinforcement Learning}\label{sec:mesc-irl}

We now describe our method for learning a set of soft constraints from a set of demonstrations $\demonstrations$ and a nominal MDP $\nominalmdp$. Our method described here generalizes the work of both \citet{scobee-2020-maximumlikelihood} and \citet{anwar-2021-icrl} to the setting of non-deterministic MDPs and soft constraints. Following \citet{ziebart-2008-maxentropy}, our goal is to optimize a function that linearly maps the features of each transition to the reward associated with that transition, $R(s_t, a_t, s_{t+1}) = \omega \cdot \phi(s_t, a_t, s_{t+1})$, where $\omega$ is the reward weight vector. \citet{ziebart-2008-maxentropy} propose a maximum entropy model for finding a unique solution ($\omega$) for this problem. Based on this model, the probability of finite-length trajectory $\tau$ being executed by an agent traversing an MDP $\mathcal{M}$ is exponentially proportional to the reward earned by that trajectory and can be approximated by $P(\tau|\omega) \approx \frac{e^{\omega^T\phi(\tau)}}{Z(\omega)} \prod_{(s_t, a_t, s_{t+1}) \in \tau} P(s_{t+1} | s_t, a_t)$.
%
The optimal solution is obtained by finding the maximum likelihood of the demonstrations $\demonstrations$ using this probability distribution, hence: $$\omega^* = \underset{\omega}{\text{argmax}} \sum_{\tau \in \mathcal{D}} \log P(\tau|\omega).$$

We extend the setting of \citet{scobee-2020-maximumlikelihood} to learning a set of \emph{soft} constraints which best explain $\demonstrations$. Allowing us to move from the notion of a constraint forbidding an action or a state to a soft constraint imposing a penalty proportional to the gravity of its violation. Given access to $\nominalmdp$ and a set of demonstrations $\demonstrations$ in ground-truth constrained MDP  $\trueconstrainedmdp$ we want to find the costs $\constraints$. Formally, we define the residual reward function $R^{\mathcal{R}}: \mathcal{S} \times \mathcal{A} \times \mathcal{S} \rightarrow \mathbb{R_+}$ as a mapping from the transitions to the penalties. We can now formally define our soft-constrained MDP $\learnedmdp$ as follows:
\begin{definition}
Given $\nominalmdp = \langle \mathcal{S}, \mathcal{A}, P, \mu, \phi, R^\mathcal{N} \rangle$ we define \emph{soft-constrained MDP} $\learnedmdp = \langle \mathcal{S}, \mathcal{A}, P, \mu, \phi, R^\mathcal{C} \rangle$ where $R^\mathcal{C} = R^{\mathcal{N}} - R^{\mathcal{R}}$. \end{definition}
\noindent
Thus, the goal of our task is to find a residual reward function $R^{\mathcal{R}}$ that maximizes the likelihood of the demonstrations $\mathcal{D}$ given the nominal MDP $\mathcal{M}^\mathcal{N}$.

Our solution is based on adapting Maximum Causal Entropy Inverse Reinforcement learning \cite{ziebart-2008-maxentropy,ziebart-2010-causalentropy} to soft-constrained MDPs. 
Following the setting of  \citet{ziebart-2008-maxentropy}  we can write the reward function $R^\mathcal{N}$ (resp. $R^\mathcal{C}$) of $\mathcal{M}^\mathcal{N}$ (resp. $\learnedmdp$) as a linear combination of the transitions:
    $R^\mathcal{N}(s_t, a_t, s_{t+1}) = \omega^\mathcal{N} \phi(s_t, a_t, s_{t+1})$ and 
    $R^\mathcal{C}(s_t, a_t, s_{t+1}) = \omega^\mathcal{C}  \phi(s_t, a_t, s_{t+1})$.
%
As, both reward functions $R^\mathcal{N}$ and $R^\mathcal{C}$ are linear, $R^\mathcal{R}$ should be linear as well: 
    $R^\mathcal{R} = \omega^\mathcal{R}\phi(s_t, a_t, s_{t+1})$.
%
From this formulation of $R^\mathcal{R}$ we can infer that the reward vectors follow 
    $\omega^\mathcal{C} = \omega^\mathcal{N} - \omega^\mathcal{R}$.

We can use Max Entropy IRL for learning a reward function compatible with the set $\demonstrations$. The gradient for maximizing the likelihood in this setting is defined as in \citet{ziebart-2008-maxentropy}:
\begin{equation*}\label{const_gradient}
    \resizebox{0.90\linewidth}{!}{
    $\nabla_{\omega^\mathcal{C}} \mathcal{L}(\mathcal{D}) = \mathbb{E}_{\mathcal{D}}[\phi(\tau)] - \sum_{(s_t, a_t, s_{t+1})} D_{s_t, a_t, s_{t+1}}  \phi(s_t, a_t, s_{t+1})$.
    }
\end{equation*}
Where $D_{s_t, a_t, s_{t+1}}$ is the expected feature frequencies for transition $(s_t, a_t, s_{t+1})$ using the current $\omega^\mathcal{C}$ weights. As the reward vectors follow $\omega^\mathcal{C} = \omega^\mathcal{N} - \omega^\mathcal{R}$, we have 
    $\nabla_{\omega^\mathcal{C}} = - \nabla_{\omega^\mathcal{R}}$.
%
Finally, by substituting this in the above, we obtain the gradient of likelihood of the constrained trajectories w.r.t. $\omega^\mathcal{R}$: 
\begin{equation*}
    \resizebox{0.90\linewidth}{!}{
    $\nabla_{\omega^\mathcal{R}} \mathcal{L}(\mathcal{D}) = \sum_{(s_t, a_t, s_{t+1})} D_{s_t, a_t, s_{t+1}}  \phi(s_t, a_t, s_{t+1}) - \mathbb{E}_{\mathcal{D}}[\phi(\tau)]$.
    }
\end{equation*}
As we estimate the residual rewards w.r.t. the nominal rewards, these rewards are automatically scaled to be compatible with the nominal rewards. 

\begin{figure*}
\begin{minipage}[t]{.33\textwidth}
  \centering
  \includegraphics[width=\linewidth]{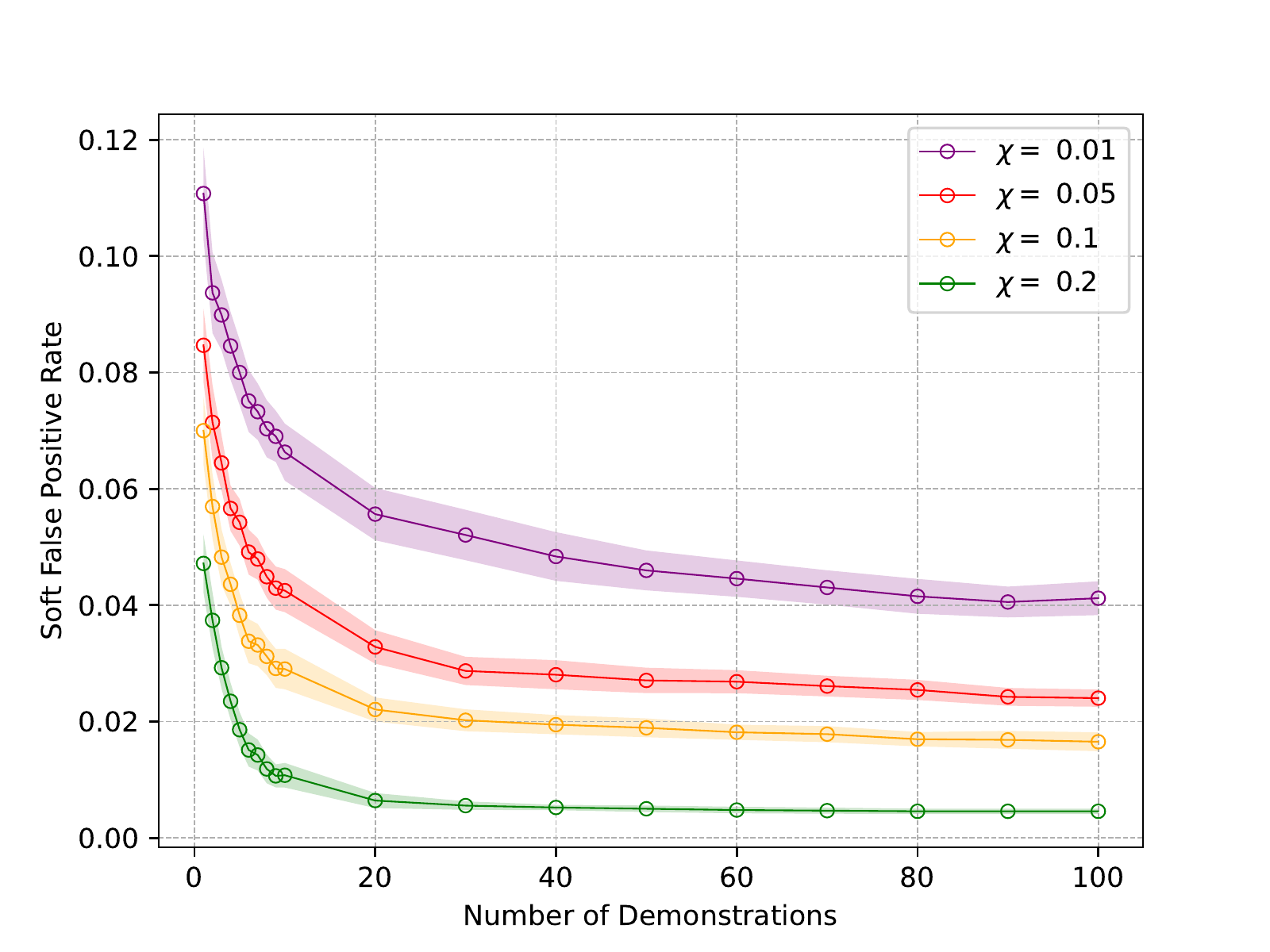}
\end{minipage}%
\hfill
\begin{minipage}[t]{.33\textwidth}
  \centering
  \includegraphics[width=\linewidth]{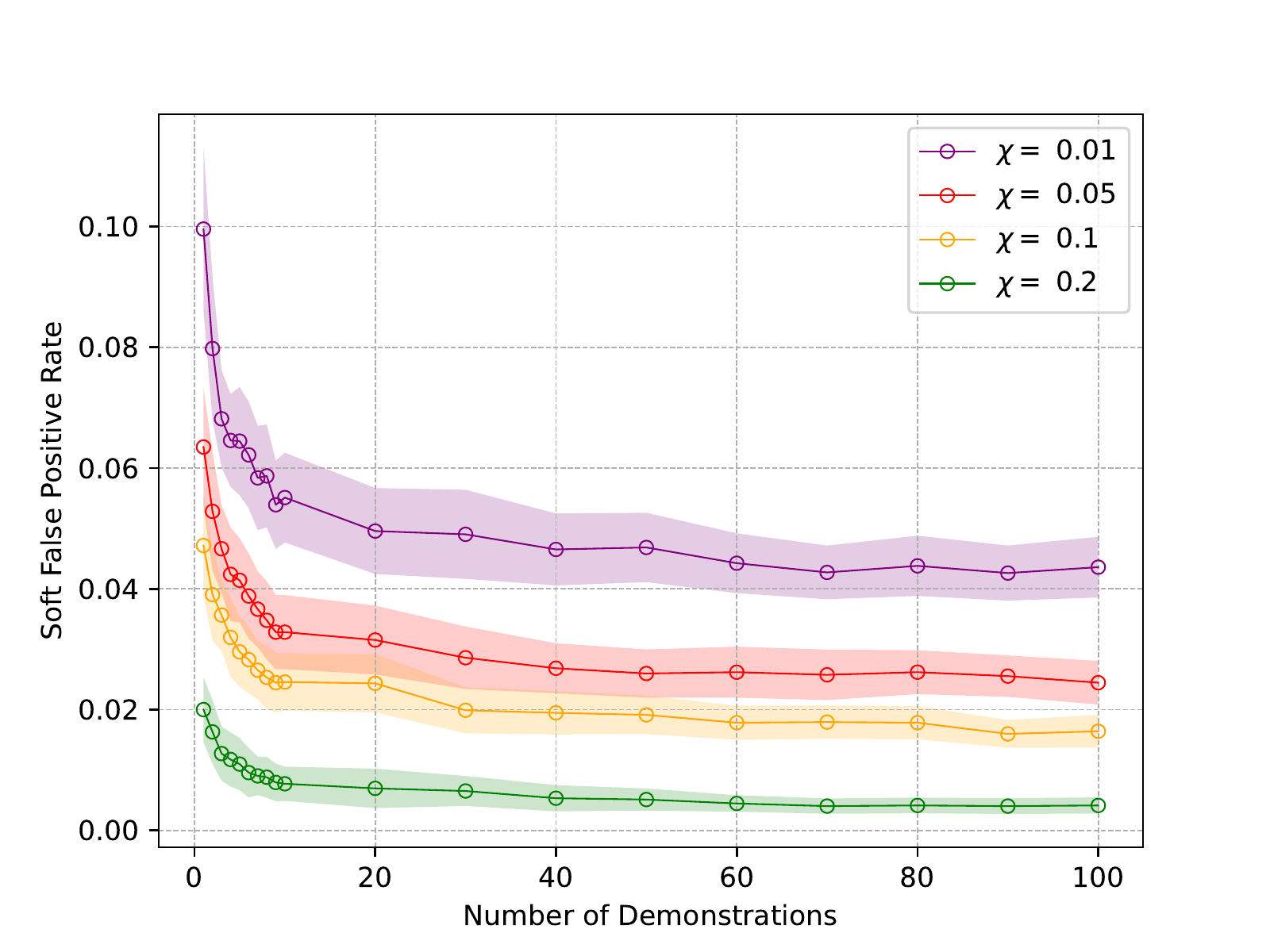}
\end{minipage}%
\hfill
\begin{minipage}[t]{.33\textwidth}
  \centering
  \includegraphics[width=\linewidth]{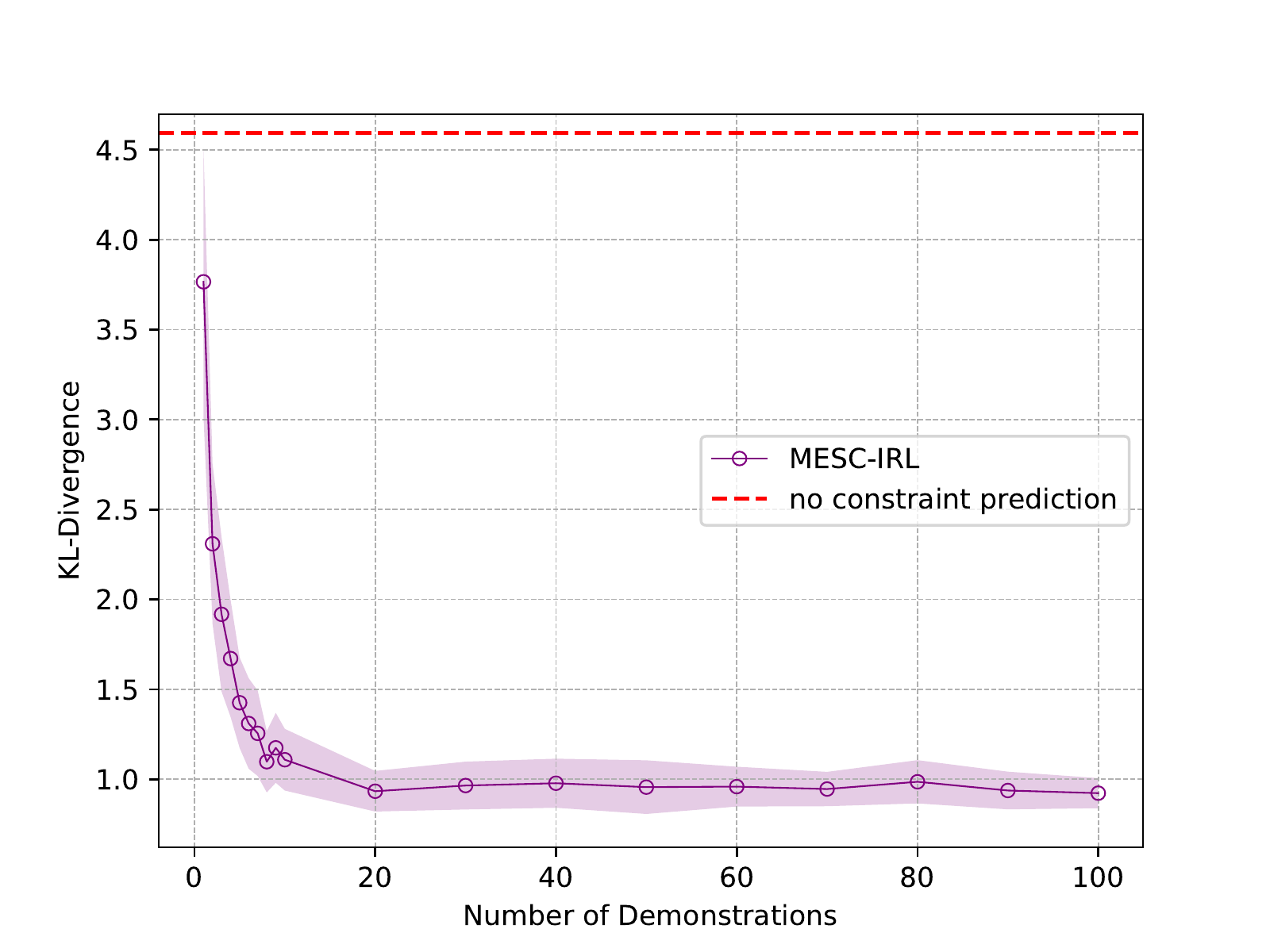}
\end{minipage}
\caption{Performance of MESC-IRL on recovering soft constraints in deterministic settings (left) and non-deterministic settings (center) according to false positive rate as well as and KL-Divergence to $\demonstrations$ for the non-deterministic setting (right). We see that across all these settings we are able to accurately recover constraints and generate behavior similar to the $\demonstrations$ even with few demonstrations.}
\label{fig:soft_compare}
\end{figure*}

\section{Generalizing From Penalties to Probabilities}\label{sec:prob}

The estimated penalties from the previous section can effectively guide an agent to navigate the environment optimally as well as provide estimates of the cost of the constraints scaled to the value of the original reward function. However, there may be instances, such as when comparing with hard constraints as we do with the work of \citet{scobee-2020-maximumlikelihood} in Section \ref{sec:emp}, where we desire \emph{probabilities} that a particular action, state, or state feature is constrained. Having probabilities allows us to: compare constraints across environments with possibly different scales, use this information to guide our policies, and to evaluate the confidence we have in a particular constraint. In this section we describe a method to transition from penalties to probabilities, as well as a generalized method to extract these probabilities based on a subset of the features of the environment, which can facilitate transfer learning between domains as described in Section \ref{sec:emp}.

A transition where the residual reward, i.e., the penalty, is significantly larger than zero is more likely to be a constraint. We estimate the significance of a penalty by scaling it to the standard deviation of the mean learned reward. Therefore, we assume that a transition penalty is a random variable, denoted by $\mathbb{C} \sim logistic(\sigma_{pooled}, \sigma_{pooled})$, following a logistic distribution with standard deviation $\sigma_{pooled}$, where $\sigma_{pooled} = \sqrt{\nicefrac{(\sigma_{\mathcal{N}}^2 + \sigma_{\mathcal{C}}^2)}{2}}$ and $\sigma_{\mathcal{N}}$ and $\sigma_{\mathcal{C}}$ are the standard deviations of the rewards in the nominal and learned constrained worlds, respectively. When penalties are close to zero, we want their probabilities to be small. To do this we set the mean of the distribution to be $\mu=\sigma_{pooled}$.

We now want to reason about a random variable $\zeta$ that indicates our belief that the transition $(s_t, a_t, s_{t+1})$ is forbidden. Using the above probability distribution we can define the probability of constraint given a transition as:
\begin{equation*}
    \resizebox{0.90\linewidth}{!}{
    $\zeta \equiv  P\left(\mathbb{C} \leq R^{R}(s_t, a_t, s_{t+1})\right) = sigmoid\left(\nicefrac{R^{R}(s_t, a_t, s_{t+1}) - \sigma_{pooled}}{\sigma_{pooled}}\right)$.
    }
\end{equation*}

In our formulation, the residual rewards only depend on the features associated with them. Hence, we can use this fact to reason about constraints over only a subset of features $\mathbf{f}$, e.g., only color or state position. Let $\phi_f \subseteq \phi$ be the subset of features we are concerned with. In our grids we represent $\phi$ with a vector of length 92. The first 81 elements represent the states, the next 8 represent the actions, and the last 3 represent the colors. So if we are interested in only learning about constraints over the colors, $\phi_f$ will be a vector equal to the last three elements of $\phi$ that is $\phi_{color} \equiv \phi_{90, 91, 92}$. 

Let $\phi_{\mathbf{f}}$ and $\omega^{R}_{\mathbf{f}}$ be the feature function and residual feature weight vector for $\mathbf{f}$. We can now define the probability of a feature value to be constrained 
as: 
\begin{equation*}
    \zeta_{\mathbf{f}} \equiv P(\mathbb{C} \leq \omega^{R}_{\mathbf{f}} \phi_{\mathbf{f}}) = sigmoid\left(\nicefrac{\omega_{\mathbf{f}} \phi_{\mathbf{f}} - std_{pooled}}{std_{pooled}}\right).
\end{equation*}


\section{Experimental Evaluation of MESC-IRL}\label{sec:emp}

In this section we empirically validate MESC-IRL for soft constraint learning against the method of \citet{scobee-2020-maximumlikelihood} for learning hard constraints in deterministic settings, an independent evaluation on learning soft constraints in non-deterministic settings, and on the ability of MESC-IRL to facilitate transfer learning between domains. Within our experiments we make use of two parameters: $\zeta$ which is a way to convert probabilities to hard constraints, i.e., it controls the probability threshold beyond which we judge a learned constraint to be a hard constraint, i.e., is only necessary to facilitate comparison to \citet{scobee-2020-maximumlikelihood}, and $\chi$ which is a threshold we use to judge whether or not the probability of a constraint is a false positive, described in more detail later. While one could tune these when deploying a system, it is not strictly necessary and they are only present here to provide evaluation metrics for comparison.

In our testing we consider only the false positive rate of our methods for two reasons. First, \citet{scobee-2020-maximumlikelihood} only consider false positives so this enables a direct coparision. Second, looking at false negatives in constraint learning can be very misleading, $\demonstrations$ may contain little or no information about states that are not relevant to the underlying path finding problem. For example, there could be a set of blocked states in the upper left of the grid, but if no demonstration and no reasonable path goes near them, then we would never know they are constrained and we would have many false negatives. Our method is inherently biased towards finding a minimal set of constraints to explain the demonstrations and hence false negative rate is not a meaningful metric.

\subsection{MESC-IRL and Hard Constraints}

Figure \ref{fig:stepping} shows the performance of MESC-IRL compared to the method proposed by \citet{scobee-2020-maximumlikelihood} on the same metrics from their paper: false positives, i.e., predicting a constraint when one does not exist, and KL-Divergence from the demonstrations set $\demonstrations$. For this test we use the same single grid, hard constraints, and a deterministic setting to allow for a direct comparison. We generate 10 independent sets of 100 demonstrations and report the mean over all these runs with the standard deviation in the shaded area. In order to decide if the values returned by MESC-IRL represent a hard constraint, we report the threshold value of $\zeta$ at various points and plot the comparison to the best result from \citet{scobee-2020-maximumlikelihood}. MESC-IRL with $\zeta \geq 0.6$, i.e., interpreting any state that has a probability of being constrained $\geq 0.6$ as a hard constraint, performs best on both false positive rate and on KL-divergence from $\demonstrations$. We observe that setting $\zeta \geq 0.7$ performs better on false positive rate but more poorly on KL-divergence as it is too selective, i.e., does not consider enough states to be constrained. In comparison to \citet{scobee-2020-maximumlikelihood}, we see that MESC-IRL actually performs on par in terms of KL-divergence. Hence we can conclude that our method is on par or better at recovering constraints and generating trajectories similar to $\demonstrations$, but is also able to work for soft constraints and non-deterministic settings.

\begin{figure*}[ht]
\begin{minipage}[t]{.32\textwidth}
  \centering
  \includegraphics[width=\linewidth]{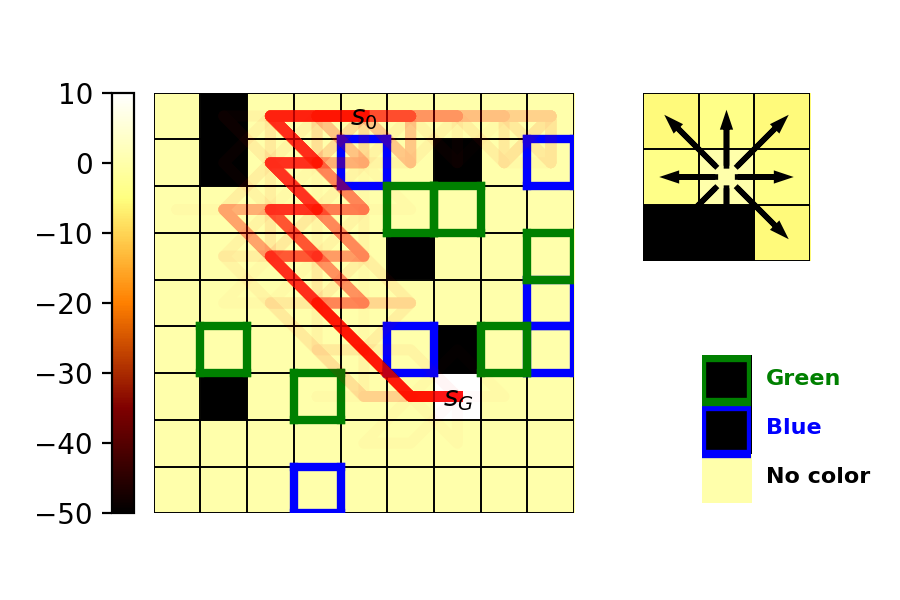}
\end{minipage}%
\hfill
\begin{minipage}[t]{.32\textwidth}
  \centering
  \includegraphics[width=\linewidth]{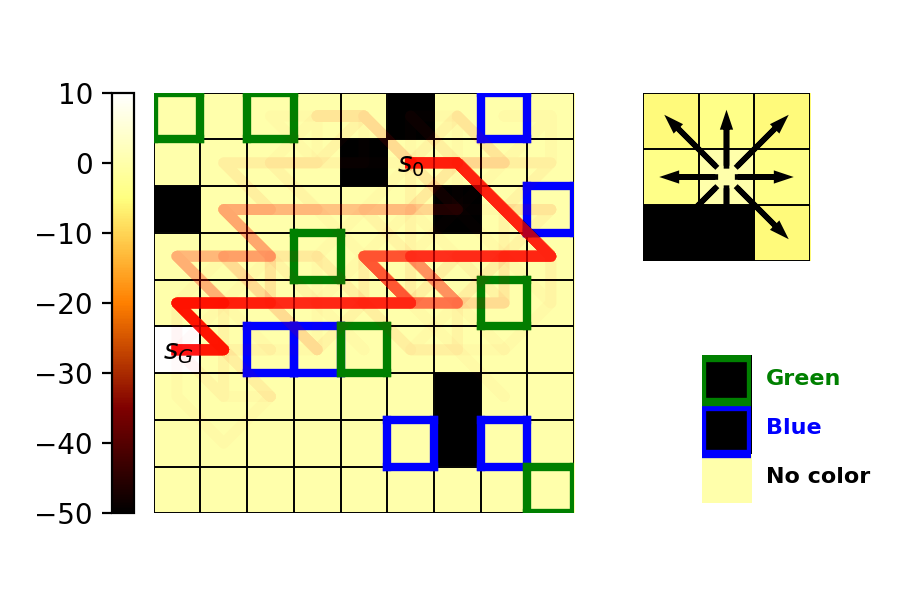}
\end{minipage}%
\hfill
\begin{minipage}[t]{.32\textwidth}
  \centering
  \includegraphics[width=\linewidth]{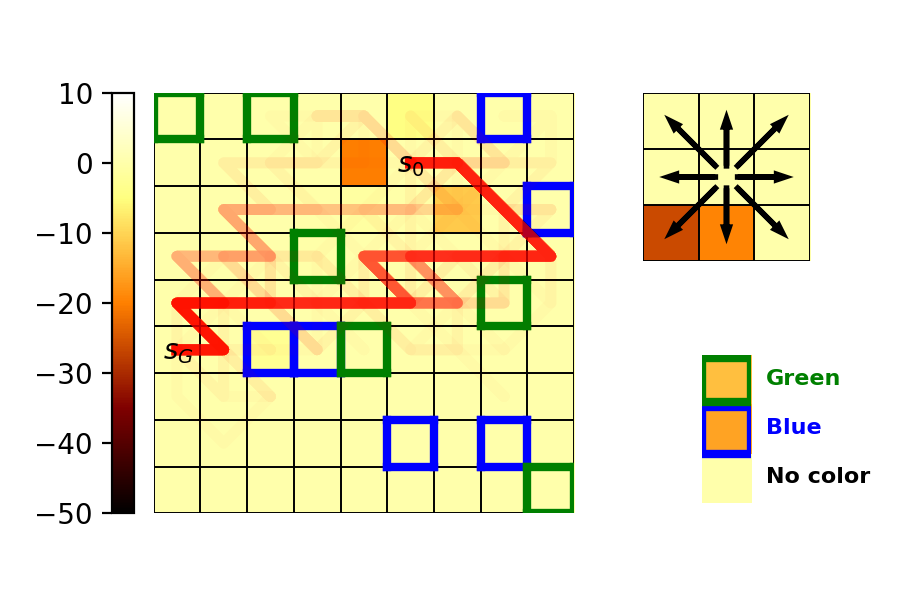}
\end{minipage}%
\caption{An example grid used for learning constraints (left). The weights learned from the left grid are transferred to the learning of the next grid (center).  A randomly generated grid that served as the ground-truth for evaluation of the transfer learning. The result of the transfer learning (right). } 
\label{fig:transfer_grid}
\end{figure*}

\begin{figure}[!h]
\centering
\begin{minipage}[t]{0.8\linewidth}
  \centering
  \includegraphics[width=\linewidth]{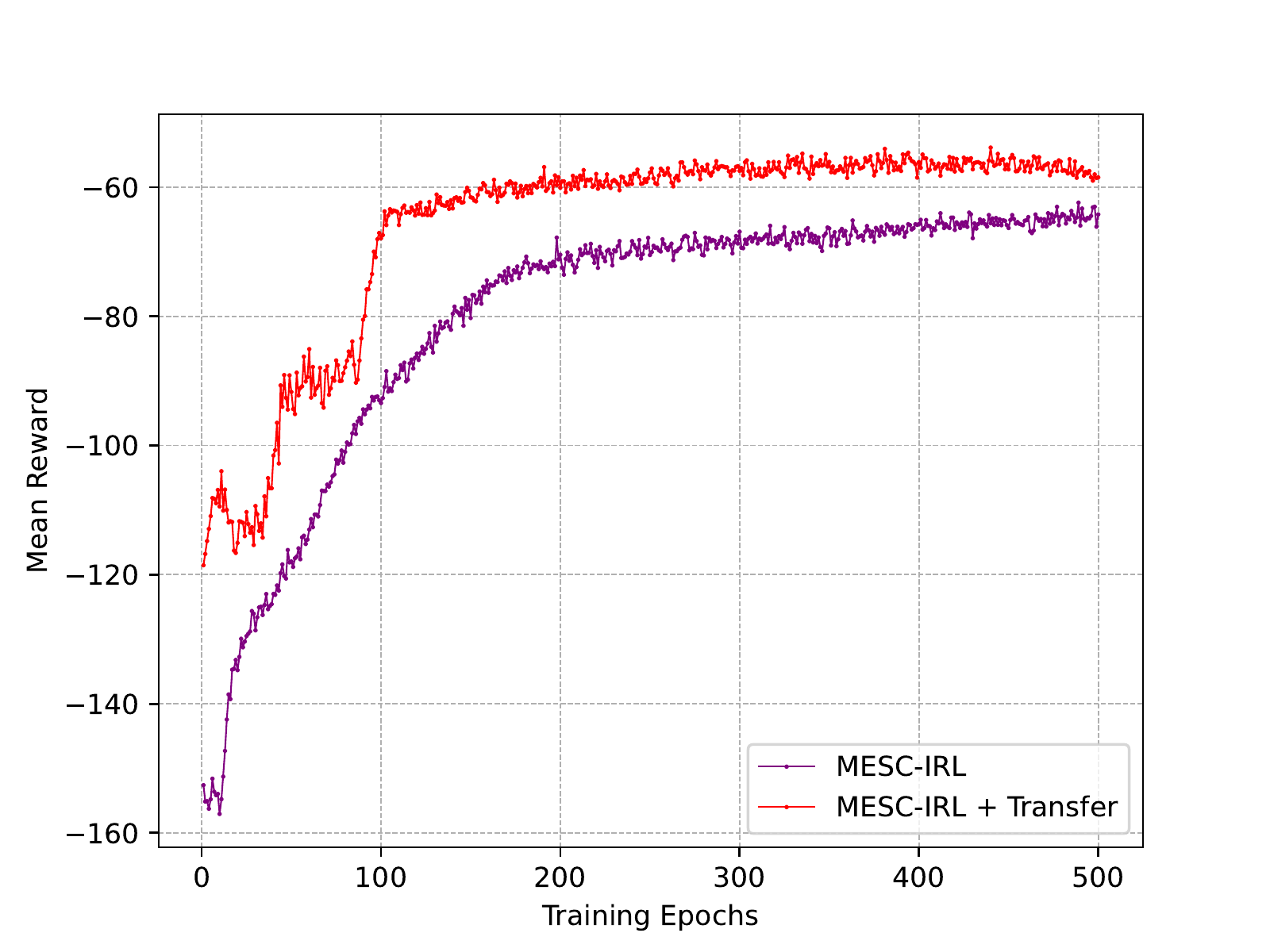}
\end{minipage}%
\hfill
\begin{minipage}[t]{0.8\linewidth}
  \centering
  \includegraphics[width=\linewidth]{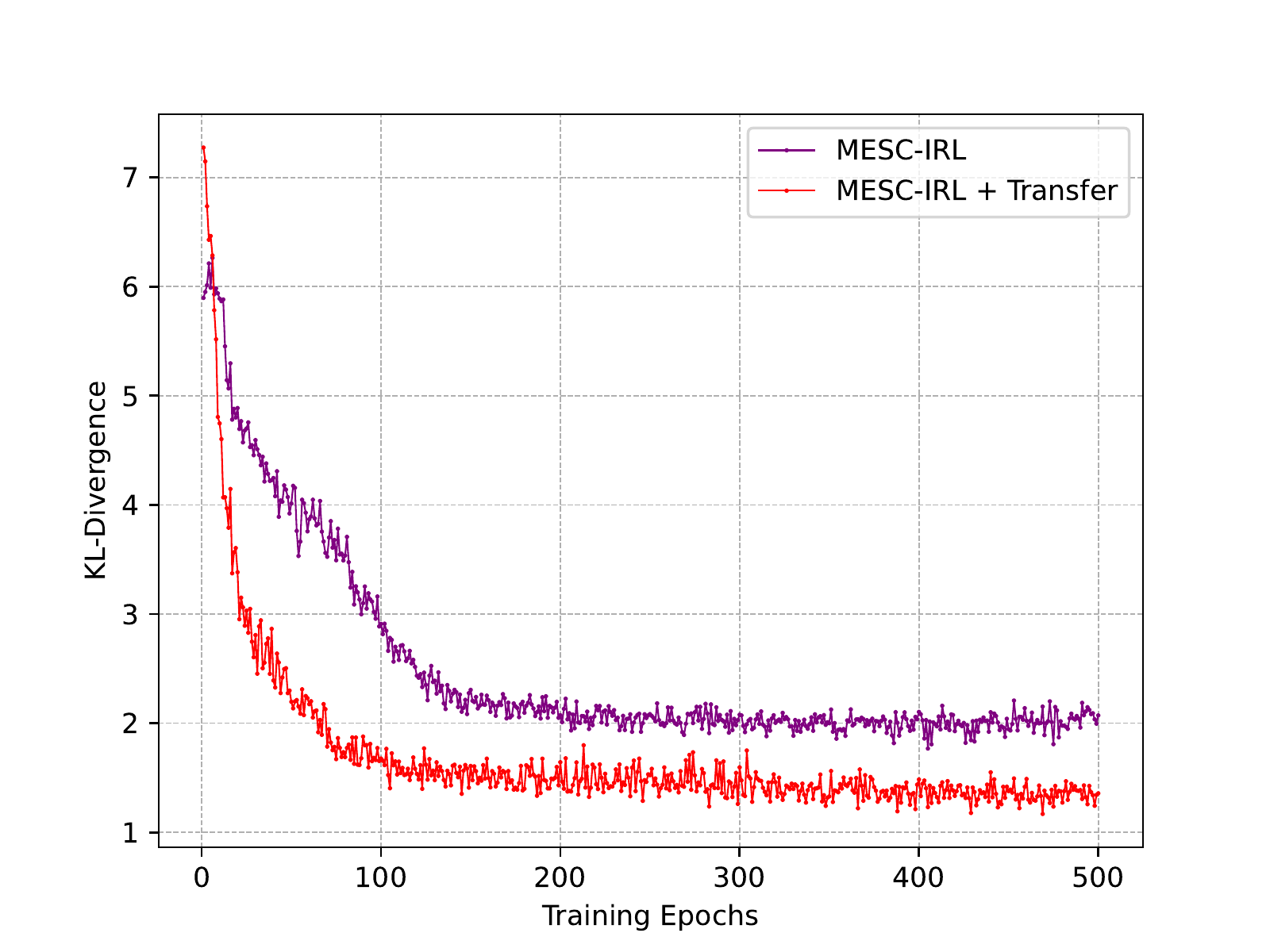}
\end{minipage}
\caption{Performance of MESC-IRL on the transfer learning task in terms of mean reward (left) and and KL-Divergence from $\demonstrations$ (right) as we vary the number of training epochs. Each point is the mean of 50 random grids.}
\label{fig:transfer}
\end{figure}

\subsection{MESC-IRL and Soft Constraints}

To evaluate MESC-IRL on soft constraints we need to adapt the notion of false positives. Let the false positive rate $fp$ be:
\begin{equation*}
    \resizebox{0.7\linewidth}{!}{
    $fp = \frac{\Big|\big\{x \in \constraints \mid \constraints^*(x) = 0 \wedge (\zeta_{\mathcal{C}}(x) - \zeta_{\mathcal{C}^*}(x) > \chi)\big\} \Big|}{\text{Num. Constraints}}$,
    }
\end{equation*}
where $\zeta_{\mathcal{C}}(x)$ and $\zeta_{\mathcal{C}^*}(x)$ are the predicted and true probability of transition $x$ being constrained as described in Section \ref{sec:prob}, and $\chi$ is a value in [0,1]. Intuitively, we count a constraint as a false positive whenever it exists in the believed constraint set $\constraints$ and there is no constraint in $\trueconstrainedmdp$ and the predicted probability exceeds the true probability by more than the threshold $\chi$. 

Figure \ref{fig:soft_compare} shows the results of our tests on recovering soft constraints in both deterministic and non-deterministic settings with random grids. For these tests we choose a start and a goal state randomly at least 8 moves apart, set 6 states for blue, 6 for green randomly, and select 6 randomly constrained states; all penalties are set to $-50$. Again we take 10 sets of 100 demonstrations. We see a strong decrease in false positive rate and KL-divergence from $\demonstrations$ as the size of $\demonstrations$ grows. We see that in general, and even more so when the threshold that minimizes the false positive rate ($\chi = 0.2$) is selected, our method almost never adds constraints that are not present in the ground truth, even for small demonstration sets. Likewise our method is able to generate trajectories very close to $\demonstrations$, showing that we are able to recover both constraints and behavior, even with soft constraints in non-deterministic setting. Hence MESC-IRL is able to work across a variety of settings and accurately capture demonstrated constraints.

\subsection{MESC-IRL and Transfer Learning}

Our final test is to see how effectively we can transfer knowledge of constraints from one grid to another in the non-deterministic setting. The goal is to see that our method is able to maintain information about, e.g., the penalty of a square being a particular color or the penalty for a particular action when we change the grid layout. To do this, we start with a base grid (e.g. Fig. \ref{fig:transfer_grid} left) and use MESC-IRL to learn a set of constraints $\constraints$ from 50 demonstrations. We then generate a new grid of the same size but vary the start and end positions, the positions of the colored squares, and the location of the constrained states (e.g. Fig. \ref{fig:transfer_grid} right). We then are given a new set of 50 demonstrations for the new grid and use MESC-IRL to learn the constraints on this new grid in two ways. First, MESC-IRL+Transfer, by transferring the learned constraints $\constraints$ from the initial grid (e.g. Fig. \ref{fig:transfer_grid} center), and second, MESC-IRL, by starting from scratch. We repeat this process 50 times and the mean reward and KL-divergence from the second demonstration set are shown in Figure \ref{fig:transfer}. We can see that the agent that has been given access to the previous learned constraints is able to perform better along both metrics, and converges faster than the agent without transfer learning. Interestingly, MESC-IRL+Transfer is able to achieve a better overall reward in the new grid by leveraging the transferred constraints, as it is able to effectively use information from both demonstration sets.

\section{Conclusion}

We have proposed a novel and general constraint learning method MESC-IRL that is able to learn soft constraints over actions, states, and state features in non-deterministic environments from a set of demonstrations; a generalization of the current state of the art. Our method provides the ability to then transfer these constraints between environments, allowing agents to retain the knowledge of constraints already learned. Important directions for future work is to expand our method to settings with continuous action spaces, possibly leveraging the work of \citet{anwar-2021-icrl}, and testing our methods in more complex environments. Another interesting direction is investigating how to relax the requirement that the reward function of $\nominalmdp$ is known, it would be interesting to explore cases where there are even unknown rewards in the base environment.

\subsection*{Acknowledgements}
Nicholas Mattei was supported by NSF Award IIS-2007955 and an IBM Faculty Research Award. K. Brent Venable are supported by NSF Award IIS-2008011.

\end{document}